\documentclass[times,twocolumn,final]{elsarticle}

\usepackage{yjvci}
\usepackage{framed,multirow}
\usepackage{amssymb}
\usepackage{latexsym}

\usepackage{url}
\usepackage{xcolor}
\usepackage{multirow}
\usepackage{hyperref}
\usepackage{graphicx}
\definecolor{newcolor}{rgb}{.8,.349,.1}

\journal{}

\begin{document}

\verso{Wang Xiaohua \textit{et~al.}}

\begin{frontmatter}

\title{Two-level Attention with Two-stage Multi-task Learning for Facial Emotion Recognition}

\author[1,2]{Wang \snm{Xiaohua}}
\author[1]{Peng \snm{Muzi}}
\author[1]{Pan \snm{Lijuan}}
\author[1]{Hu \snm{Min}\corref{cor1}}
\cortext[cor1]{Corresponding author}
\ead{jsjxhumin@hfut.edu.cn}
\author[2]{Jin \snm{Chunhua}}
\author[1,3]{Ren \snm{Fuji}}

\address[1]{Hefei University of Technology, Hefei, China}
\address[2]{Huaiyin Institute of Technology, Huaian, China}
\address[3]{University of Tokushima, Tokushima, Japan}

\received{*}
\finalform{*}
\accepted{*}
\availableonline{*}
\communicated{**}

\begin{abstract}
Compared with facial emotion recognition on categorical model, the dimensional emotion recognition can describe numerous emotions of the real world more accurately. Most prior works of dimensional emotion estimation only considered laboratory data and used video, speech or other multi-modal features. The effect of these methods applied on static images in the real world is unknown. In this paper, a two-level attention with two-stage multi-task learning (2Att-2Mt) framework is proposed for facial emotion estimation on only static images. Firstly, the features of corresponding region(position-level features) are extracted and enhanced automatically by first-level attention mechanism. In the following, we utilize Bi-directional Recurrent Neural Network(Bi-RNN) with self-attention(second-level attention) to make full use of the relationship features of different layers(layer-level features) adaptively. Owing to the inherent complexity of dimensional emotion recognition, we propose a two-stage multi-task learning structure to exploited categorical representations to ameliorate the dimensional representations and estimate valence and arousal simultaneously in view of the correlation of the two targets. The quantitative results conducted on AffectNet dataset show significant advancement on Concordance Correlation Coefficient(CCC) and Root Mean Square Error(RMSE), illustrating the superiority of the proposed framework. Besides, extensive comparative experiments have also fully demonstrated the effectiveness of different components.

\end{abstract}


\end{frontmatter}

\section{Introduction}
\label{sec1}
Emotion recognition has been a hot topic in the field of computer vision, and attracted much attention in recent years. As a significant component of Human-Computer Interation(HCI), an outstanding emotion recognition framework is a crucial role. The perception of human emotions can advance the degree of humanization, and the usability of many applications, such as elderly escort robots, online education, interactive games, etc.

As the one of most powerful and natural signals of expressing emotion states~\cite{darwin1998expression}, facial emotions account for the 55\% role of emotional information~\cite{mehrabian2008communication}. Due to the influence of many factors, such as different subjects, races, illumination, complex background and so on, facial emotion analysis is a indubitable challenging task. Most of the previous researches~\cite{shan2009facial,liu2012facial,lopes2017facial} were based on data in laboratory-controlled environment which can avoid many factors mentioned above with the limitation of the number of datasets and algorithm. Compared to facial emotion recognition(FER) in laboratory-controlled environment, FER in-the-wild is closer to the business application. Nowadays, several works about FER in-the-wild has gradually emerged~\cite{guo2016deep,ding2017facenet2expnet}. On the one hand, with the rapid development of network, countless images with facial emotion are posted in the social media every second; on the other hand, the rapid growth of computing power makes it possible for deep learning to fit a robust model. However, those non-emotion factors still can't be properly solved although recent advances in neural network(NN) have supposed momentous development in this field. Therefore, it is critical to diminish the impact of non-emotion factors. Many prior studies used Convolutional Neural Networks(CNN) to learn features related to emotion through metric learning. Meng~\cite{meng2017identity} proposed an identity-sensitive contrastive loss to learn identity-related information from identity labels to diminish the burden brought by different subjects. Cai~\cite{cai2018island} et al. designed a novel island loss which is defined as the summation of center loss~\cite{wen2016discriminative} and distances between different categories of feature centers. The center loss diminish intra-class distance while the distances between different categories amplify inter-class distance. These methods based on metric learning could improve the discriminative ability but not for continuous models. To curtail the harmful influence of non-emotion factors, Sun~\cite{sun2018visual} introduced attention mechanism to second to last fully connected layer for feature mapping. The author claimed that the approach extracted features of region of interest (ROI) automatically. However, only the high-level features were taken into account but the low-level features and the association of these two kinds of features were neglected. Therefore, a two-level attention approach is designed to overcome the problems mentioned above.

In addition, there are three main models to describe facial emotion, categorical model, facial action coding system(FACS) model and dimensional model. Ekman~\cite{ekman1971constants} defined six basic emotions(happiness, anger, sadness, surprise, disgust, fear) which represent different categories. After that, Ekman proposed FACS model, which utilized specific action units indicate facial movements to describe human affect states. On dimensional model, the value represents a continuous emotional scale which can expresses emotion more accurately, and reflects the relationship of different emotions compared with categorical model. And the facial emotion with slight changes can also be reflected by continuous numerical value effectively. Everything has two sides. The categorial model is easier to train while continuous model is more challenging. To leverage the advantage of categorical representations, we append it to continuous representations to estimate valence and arousal.

Previous works mainly focused on category model, extracted features included traditional method and deep learning were applied to the appropriate classifier or ensemble method to achieve results~\cite{shan2009facial,jung2015joint,kim2016fusing}. Recently, there have been some researches on continuous model, but most of them based on physiological signals, video, speech or multi-modal features~\cite{brady2016multi,xia2017multi,chen2017multimodal}. In aforementioned work, Long Short Time Memory(LSTM) network or Support Vector Regression(SVR) is exploited to predict the value with the usage of temporal information. However, physiological signals, video and multi-modal information are arduous to be collected, while the static image is at your fingertips. In this context, we focus on emotion Valence-Arousal model built by Russel~\cite{russell1980circumplex} for imaged-based emotion estimation. All basic emotions Ekman designed can be mapped into the Valence-Arousal model. Where valence defines the emotion intensity of positive or negative, arousal represents how exciting or calming the emotion is.  The prediction of both is essentially a regression problem. As far as the research we have discovered, Mollahosseini~\cite{mollahosseiniaffectnet} replaced the last fully connection(FC) layer with linear regression and trained arousal and valence task respectively. Nonetheless, it overlooked the correlation between the two tasks and expanded the time of train notably. Multi-task learning can utilize the relationship of tasks adequately to improve the learning of other task through the shared features of different tasks. For the inherent relation of two tasks(valence and arousal) and different representations mentioned above(categorical and dimensional), a two-stage multi-task learning structure is proposed in the following research.

In this paper, we propose a two-level attention with two-stage multi-task learning(2Att-2Mt) framework to estimate valence and arousal simultaneously. At first, the residual attention mechanism proposed by Wang~\cite{wang2017residual} is adopted to extract the features of different layers as the first-level attention, which introduce attention mechanism to different level layers. Then, the convolution layer with 1*1 filter and global average pooling layer (GAP) are utilized to transform features extracted by 1st-level attention. In the following, we propose to use Bidirectional-Recurrent Neural Network (Bi-RNN) with self-attention to capitalize the information of different layers with diverse receptive fields. In other word, 2nd-level attention distilled the information of relationship of global and local features automatically. In addition, we implement multi-task learning on our framework. On the one hand, we utilize multi-task learning to learn valence-arousal and category representations to leverage the benefits of the two; on the other hand, considering the high correlation of valence and arousal, we estimate valence and arousal simultaneously through multi-task learning. To show the benefit of our proposed network, we have conducted extensive experiments on a open dataset (AffectNet~\cite{mollahosseiniaffectnet} and achieved a remarkable result.

The structure of this paper is as follow. In the next section, we cover the related work of this paper. Section~\ref{sec3} describe the network we designed in detail. Then, the experimental results and exhaustive analysis are presented in Section~\ref{sec4}. Finally, we will conclude the paper in Section~\ref{sec5}.

\section{Related work}
\label{sec2}
In this section, we will briefly introduce the related work for facial emotion recognition, attention mechanism and multi-task learning.

\subsection{Facial emotion recognition }
Extracting features that characterize facial emotion state is the most significant procedure of facial emotion recognition, whether under categorical model or continuous model. Traditional methods capture facial information (geometry, appearance and texture information) as features, i.e. Local Binary Patterns (LBP)~\cite{shan2009facial}, Histogram of Oriented Gradients (HOG)~\cite{dalal2005histograms} and facial landmarks~\cite{cootes2001active}. These hand-crafted features have been implemented on facial emotion recognition under  laboratory-controlled environment successfully. However, performance of these hand-crafted features degenerates on in-the-wild dataset with large variations in background, subject, pose, illumination, etc. The above feature extraction methods cannot deal with such non-emotion factors well, and the generalization ability under the environment is relatively impaired. In contrast, with the growth of computing power, features which are more suitable for the real world can be learned by CNN-based deep learning approaches~\cite{dhall2015video}. In virtue of the data-driven technology, the challenges brought about by non-emotion factors in the real world can be solved to a certain extent. Liu~\cite{liu2017adaptive} proposed (N+M)-tuple loss to calculate the distance between positive and negative samples to diminish the burden brought by different subjects. Li~\cite{li2017reliable} designed Deep Locality Preserving loss to preserve the compactness of intra-class samples and improve the emotion-related representation capability. Yao~\cite{yao2016holonet} uniquely designed residual structure with concatenated rectified linear unit (CReLU) and an inception-residual block for emotion recognition. Those enhance the discriminative ability of learned features. In addition, some researchers~\cite{zhang2017facial,fan2018multi} assembled multiple networks to improve the result of facial emotion recognition.

At present, most of the aforementioned approaches focused on the discrete model. Facial emotion estimation on continuous model still is a troublesome task. There have been some competitions based on continuous models, i.e. Audio/Visual Emotion Challenges(AVEC), One-Minute Gradual-Emotion Behavior Challenge (OMG-Emotion), Affect-in-the-wild Challenge(Aff-wild). In AVEC2017, the winner~\cite{chen2017multimodal} combined the features of text, acoustic and video and utilized LSTM to extract time information for the final prediction. Peng~\cite{peng2018deep} et al. joint trained audio and video model incorporated Bi-LSTM and temporal pooling together, and got the first prize in OMG-Emotion. Kollias~\cite{kollias2018multi} put the final convolution layer, pooling layer, and fully connected layer into the gated recurrent unit and fused the final results. Chang~\cite{chang2017fatauva} proposed an integrated network to extract face attribute action unit(AU) information and estimate Valence-Arousal values simultaneously and achieved the first place in Aff-wild.

For image-based facial emotion recognition, in addition to the method proposed by Mollahosseini~\cite{mollahosseiniaffectnet} which we mentioned in Section~\ref{sec1}. Zhou~\cite{zhou2018fine} also conducted the experiments on the FER-2013 dataset. The original dataset is marked as seven categories, and the author labelled these images as dimensional values by crowd-sourcing. Zhou replaced the last layer of VGG16 and ResNet with GAP layer and adopted bilinear pooling to predict the value of valence and arousal, and he achieved positive results. In this paper, we also focus on image-based facial emotion recognition.

\subsection{Attention mechanism}
The process of human perception proves the importance of the attention mechanism~\cite{mnih2014recurrent}. Broadly, attention can be seen as a mechanism for allocating available processing resources to the most signal-fertile components~\cite{hu2017squeeze}. Presently, attention mechanism is extensively used in various fields, machine translation, visual question answering, image caption. In Previous researches, most of the attention mechanisms were implemented on sequence processing. Hu~\cite{hu2017squeeze} designed Squeeze-and-Excitation Network to learn the weight of feature map in line with loss, which makes the quality features can be improved, futile features can be diluted. It can be seen as engaging attention mechanisms to feature maps on channel dimension. Qin~\cite{qin2018visual} employed attention mechanism to generate saliency maps, which have a strong positive correlation with facial emotion. The maps can be seen as learned features. Wang~\cite{wang2017residual} proposed an attention model for image classification which uses an hourglass model to construct trunk and mask branch, where mask branch is a Bottom-up Top-down structure. The mask branch is able to generate soft attention weight which corresponds to each pixel of the original input. In this paper, we exploit the residual attention block proposed by Wang to extract the features as the first-level attention.

\subsection{Multi-task learning}
Multi-task learning is widely used in computer vision and natural language processing~\cite{duan2018novel,chang2017learning}. All of the papers mentioned above showed that multi-task learning can train a unified system to carry out multiple tasks simultaneously (The premise is that there is a certain correlation between tasks). In ~\cite{xia2017multi}, the author proposed a multi-task model to adapt Deep Belief Network(DBN), in which the classification of emotion was regarded as the main task, and the prediction of valence and arousal(V-A) was regarded as the secondary task. The results showed that multi-task learning utilized the additional information between different tasks and advanced the result of emotion estimation. In~\cite{chang2017fatauva}, the author constructed a end-end network structure and trained facial attribute recognition, AU recognition and V-A estimation synchronously. Similarly, Pons~\cite{pons2018multi} et al. proposed a novel selective joint multi-task(SJMT) approach to exploit categorical and AU representation, which address multi-domain problem in multi-task learning. In this paper, we exploit a multi-task learning structure to extract the shared features representation between tasks in view of the correlation of the valence and arousal. In Kervadec's~\cite{kervadec2018cake} method, the values of valence and arousal are used to assist in predicting discrete emotions.  Inspired by Kervadec, we learned the categorial representations to compensate for inaccuracy of continuous representations and predicted the values of valence and arousal simultaneously through our proposed two-stage multi-task learning structure.

\section{Proposed framework}
\label{sec3}
We propose a novel framework for facial emotion recognition, mainly including two-level attention mechanism and two-stage multi-task learning. Two-level attention mechanism can extract emotion-related features and diminish the influence brought by non-emotion-related information automatically. Two-stage multi-task learning advance the accuracy of dimensional representations with heterogeneously representations first, and then predict the value of valence and arousal simultaneously with second-stage multi-task learning. In this section, we will introduce the overall framework, two-level attention mechanism, two-stage multi-task learning structure and regression approach.

\begin{figure*}
\centering
\includegraphics[width=0.7\textwidth]{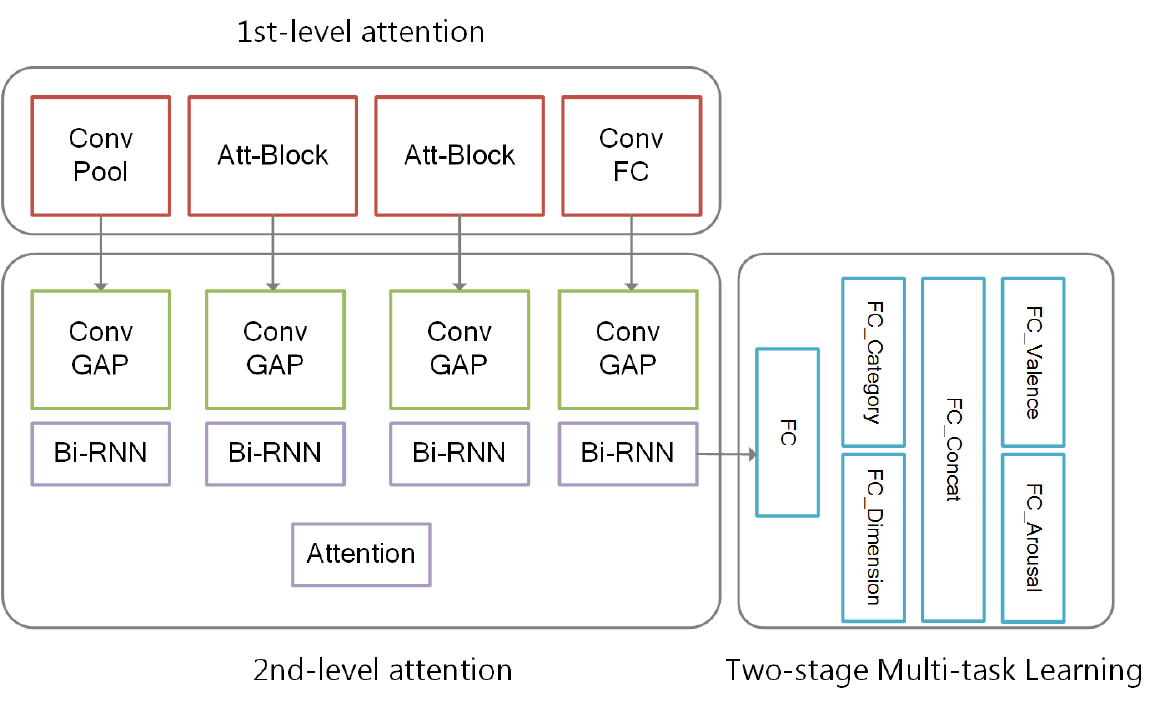}
\caption{An overview of our proposed framework. The framework is mainly contains three parts (first-level attention, second-level attention and two-stage multi-task learning).} \label{fig1}
\end{figure*}

Figure~\ref{fig1} shows the overall framework for facial emotion estimation in this paper. The proposed framework mainly consists of four parts. First, as the 1st-level attention, the position-level features of different receptive fields with attention are extracted by residual attention block~\cite{wang2017residual}. And then distill the information of relationship between different receptive field features by Bi-RNN with self-attention mechanism as the 2nd-level attention.The two-level attention mechanism not only adaptively extracts high-quality position-level features, but also fuses the information of different layer-level features. In the following, we construct an auxiliary structure with multi-task learning to capture categorial representations to complement dimensional representations we learned and estimate our targets together. The two-stage multi-task learning structure takes advantage of the strengths of the dimensional model and categorical model respectively and take into account the highly correlated information between valence and arousal. Finally, a robust regression approach is utilized to predict the values of valence and arousal.

\subsection{First-level attention}
One of the most challenging difficulties in facial emotion recognition in the real world is to weaken the influence of non-emotion factors and learn more discriminative emotion-related features. Most prior studies~\cite{liu2017adaptive, li2017reliable} have focused on reducing intra-class distance to reach the target, which could not be used for continuous models. In consequence, we utilize the residual attention block based on spatial attention to extract features for different positions automatically. It can be seen as the first-level attention features (position-level features) of this paper.

\begin{figure}
\centering
\includegraphics[width=0.45\textwidth]{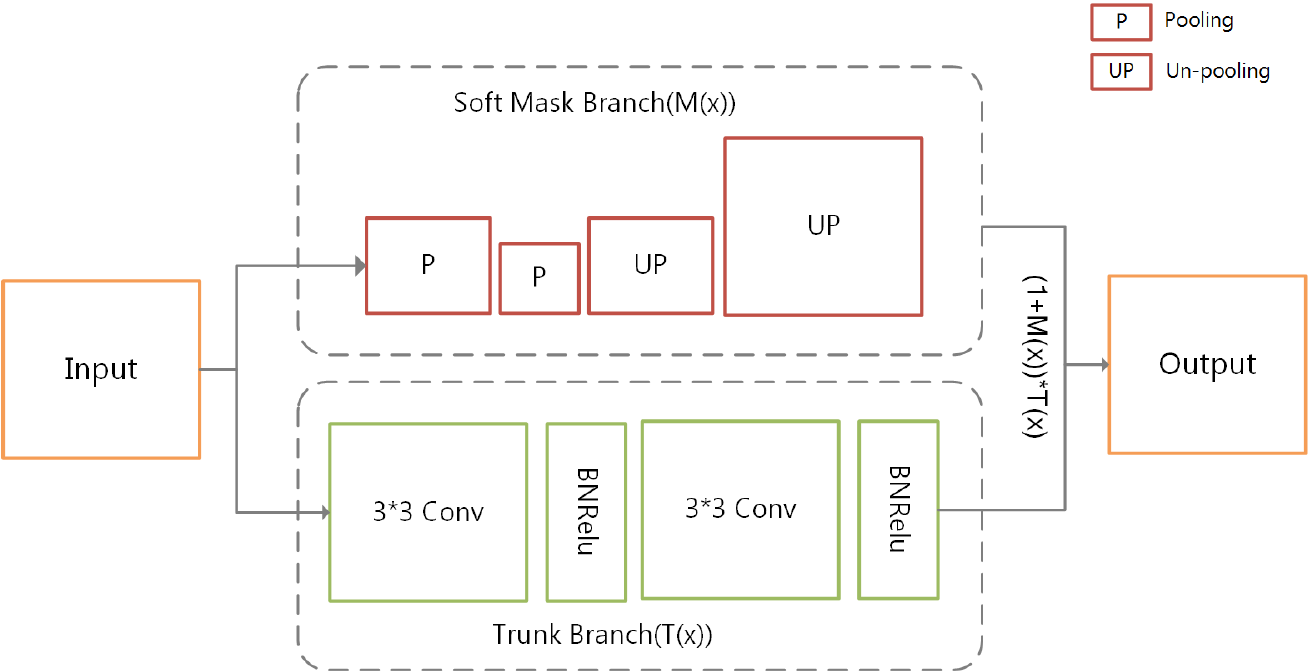}
\caption{The composition of attention block. The two branches were split from input
and merged together finally.} \label{fig2}
\end{figure}

The residual attention block adopts a bottom-up top-down structure to expand feed-forward and feedback attention process into a single feed-forward process, which makes it accessible to embed it into any end-to-end framework. As shown in the Figure~\ref{fig2}, the block has two branches: trunk branch and mask branch. The trunk branch has the same structure as the normal CNN and extracts the features by convolutional layers. The key component of the block is mask branch, which can generate soft weight attention. The mask branch utilizes several max pooling layers to expand the receptive field and obtains global information of the block. In the following, the same number of up-sample layers are manipulated to amplify feature map to the same size of input through the symmetric top-down structure (bilinear interpolation is adopted in up-sample layer). Finally, the sigmoid function is employed to regulate the output value to a range from 0 to 1, which is used as a control gate for trunk branch features. At this point, the scale of output of mask branch is the same as that of trunk branch and both can be calculated directly.
The common attention model is calculated by dot product, but in the deepening network, the features will decay incredibly since the soft attention weight less than 1. Further, the potential error of soft attention weight may undermine the favorable features that trunk branch has learned. Therefore, the block constructs an approximate identical mapping structure to settle the above problems. The output of the entire block is as follow:
\begin{equation}\label{1}
H_{i,c}(x)=(1+M_{i,c}(x))*T_{i,c}(x).
\end{equation}
Where $i$ denotes the $i^{th}$ pixel while $c$ represents the $c^{th}$ channel of the feature maps. When $M_{i,c}(x)$ is close to zero, $H_{i,c}(x)$ is basically approximate to that of $F_{i,c}(x)$. $M(x)$ is capable to enhance the high quality features and weaken the non-significant features. In the forward process, it works as the feature selector with attention, and can easily update the gradient in the backward propagation. The gradient derivative formula is as follow:
\begin{equation}\label{2}
\frac{\partial H(x;\theta,\phi)}{\partial \phi}=(1+M(x;\theta))\frac{\partial T(x;\phi)}{\partial \phi}.
\end{equation}
Where $\phi$ is the parameter of trunk branch and $\theta$ is the parameter of mask branch. The presence of mask branch enhances the robustness of the network and prevents wrong gradients from noisy label to update the trunk branch parameters. Besides, the "1" in Equation.1 enables trunk features to bypass soft mask branch and reach the output of block directly, which weakens the ability of feature selector of mask branch. Consequently, we can retain high quality features and suppress even discard some poor features appropriately through the block. In other word, we can learn the position-level of features for every block, which helps to learn the emotion-related features.

\subsection{Second-level attention}
To simulate the way human observe real objects, we could not look at the whole image simply. Firstly, most grasp the target from a global perspective, and observe it in global to local order with prior impression. In other words, people generally look at the outline and then perceive the details. Finally, people combine global and local information to comprehend and judge the target. Abstractly, the way people look at images can also be seen as a sequence model, not only the order from global to local, but also combine global with local information which corresponding to different level features. The residual attention block mentioned in the previous part only extracted features for a certain size of receptive fields. Considered the above cognition, we judiciously adopted Bi-RNN to imitate this process. We exploited the self-attention mechanism to learn the features extracted from different receptive fields adequately. Merged with Bi-RNN, we assimilated the information both in a order from global to local and local to global. Therefore, Bi-RNN combined with self-attention model was utilized to extract features of different levels.

For the sequence to sequence problem, the input of Bi-RNN is vectors of multiple time. While the input in this paper is feature vectors of different levels (different receptive fields). In generally, most prior researches used the fully connected layer as the input vectors. Considered the huge amount of parameters in FC, we designed a simple structure as illustrated in Figure~\ref{fig1}. A convolution layer with 1*1 filter was used to reduce or elevate dimensions to a fixed number for different layers. Followed by a GAP layer, the dimensions of output at each level were consistent. The input vectors can be represented as $X=(x_1,x_2,...,x_l)$. We concatenate the hidden states (forward and backward process) of Bi-RNN to vectors $H=(h_1,h_2,...,h_l)$. And the output of Bi-RNN can be represented as $Y=(y_1,y_2...,y_l)$. Therefore, the output with self-attention can be calculated:
\begin{equation}\label{3}
Y_{att}=\sum_{i=1}^{l}y_i\frac{exp(score(h_i,y_i))}{\sum_{i=1}^{l}exp(score(h_i,y_i))}.
\end{equation}
We implement dot production as the score function here. This allows the correlation of hierarchical features(low, mid, high-level) to be distilled through the self attention model. The output vectors are utilized as the input of two-stage multi-task learning in the next part.

\subsection{Two-stage multi-task learning structure}
Multi-task learning utilizes the correlation between multi-tasks and learns the shared feature representations of them. It raises the generalization ability of the model and shorten the training time tremendously. In this paper, we propose a two-stage multi-task learning structure to take advantage of the relationship between different models (categorical and dimensional) representations and different predicting targets(valence and arousal). Inspired by Kervadec~\cite{kervadec2018cake}, we employ multi-task learning to learn categorical and dimensional representations, since there is a certain correlation between the two. Hence, the first-stage multi-task learning ensures that model learns diverse characteristics which can complement each other. The "AVk" approach Kervadec~\cite{kervadec2018cake} proposed to predict discrete emotions uses only two values(valence and arousal) and did not update the parameters of continuous model. Relatively, we use a whole layer of neurons of the respective model as features rather than two values and optimized their weights simultaneously, which make greater use of the relationship between different models representations. Then, the second-stage multi-tasking learning is adopted to estimate valence and arousal simultaneously due to the high correlation between them. For multi-task learning, choosing which parts of framework as the shared layers will lead to different effects. Excessive shared layers cannot reflect the distinction between tasks, but few shared layers cannot learn the commonality of tasks. While appropriate shared layers maximize the use of the correlation of tasks and make them have certain independence.

In our proposed framework, 1st-stage multi-task learning is adopted after the feature extraction of the 2nd-level attention, and concatenate the two kinds of features. The concatenate features fuse dimensional and discrete features In the following, 2nd-stage multi-task learning is applied to predict the targets. In Section~\ref{sec4}, the experiments with multi-task learning using different strategies are also carried out to demonstrate the effectiveness of our approach. The training goal of our framework is to minimize:
\begin{equation}\label{4}
L_{total}=\alpha{L_{clf}} + (1-\alpha)(\beta L_{arousal}+(1-\beta){L_{valence}}).
\end{equation}
Where $L_{clf}$ denotes the loss of categorical representations, $\alpha$ and $\beta$ are hyper parameters for balancing two tasks in two-stage multi-task learning. In our framework, they are set to 0.5 and 0.3, respectively. The hyper parameters are determined by random search method.

\subsection{Regression method}
\begin{figure}
\centering
\includegraphics[width=0.45\textwidth]{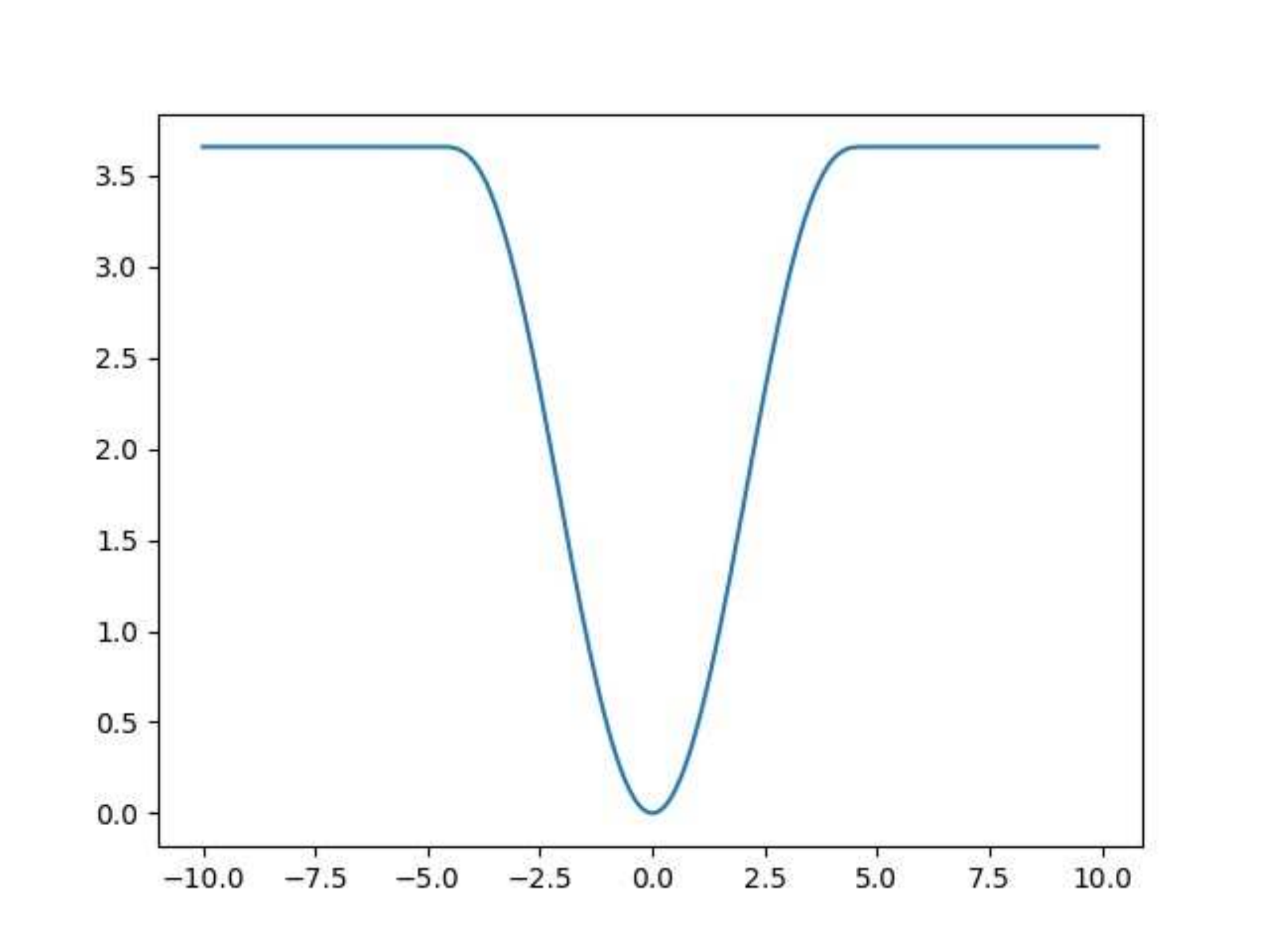}
\caption{The curve of Tukey's biweight function.} \label{fig3}
\end{figure}

In generally, mean square error(MSE) is used as the loss function in the training phase for regression model. The calculate formula of MSE is as follow:
\begin{equation}\label{5}
L=\sum_{i=1}^{n}(y_i-\hat{y_i})^2.
\end{equation}
Where $y_i$ represents the ground label and $\hat{y_i}$ represents the predicted label. Although MSE gives more punishment to samples with large error, it is sensitive to the outliers. The dataset (described in Section~\ref{sec41} was constructed by crowdsourcing method. Owing to the subjective consciousness of annotators, the dataset may contains some inconsistent or imprecise labels. Meanwhile, there are also some dirty data that could not be properly preprocessed. In this paper, we adopt Tukey's biweight function~\cite{black1996unification} as the loss function of our framework to overcome the problem. And the specific formula is as follow:
\begin{equation}\label{6}
L_{tukey}=\left\{\begin{array}{cc}
\frac{c^2}{6} * \left[ 1-(1-\frac{y_i-\hat{y_i}}{c}^2)^3 \right] & if|y_i-\hat{y_i}|\leq c \\
\frac{c^2}{6} & otherwise
\end{array}\right.
\end{equation}
Where c is a hyper parameter. It is set to 4.685 empirically. Unlike MSE, Tukey's biweight function is a non-convex function. The magnitude gradient of the noisy label sample can be reduced close to zero during back-propagation and the problem of human-labeled can be deal with effectively. The curve of Tukey's biweight function is as shown in Figure~\ref{fig3}.
\section{Experiments}
\label{sec4}
We assess the effectiveness of our proposed method for facial emotion recognition on a widely used dataset(AffectNet). The following shows the experimental results and detailed analysis.

\subsection{Dataset and Performance Measures}
\label{sec41}
\begin{figure}
\centering
\includegraphics[width=0.4\textwidth]{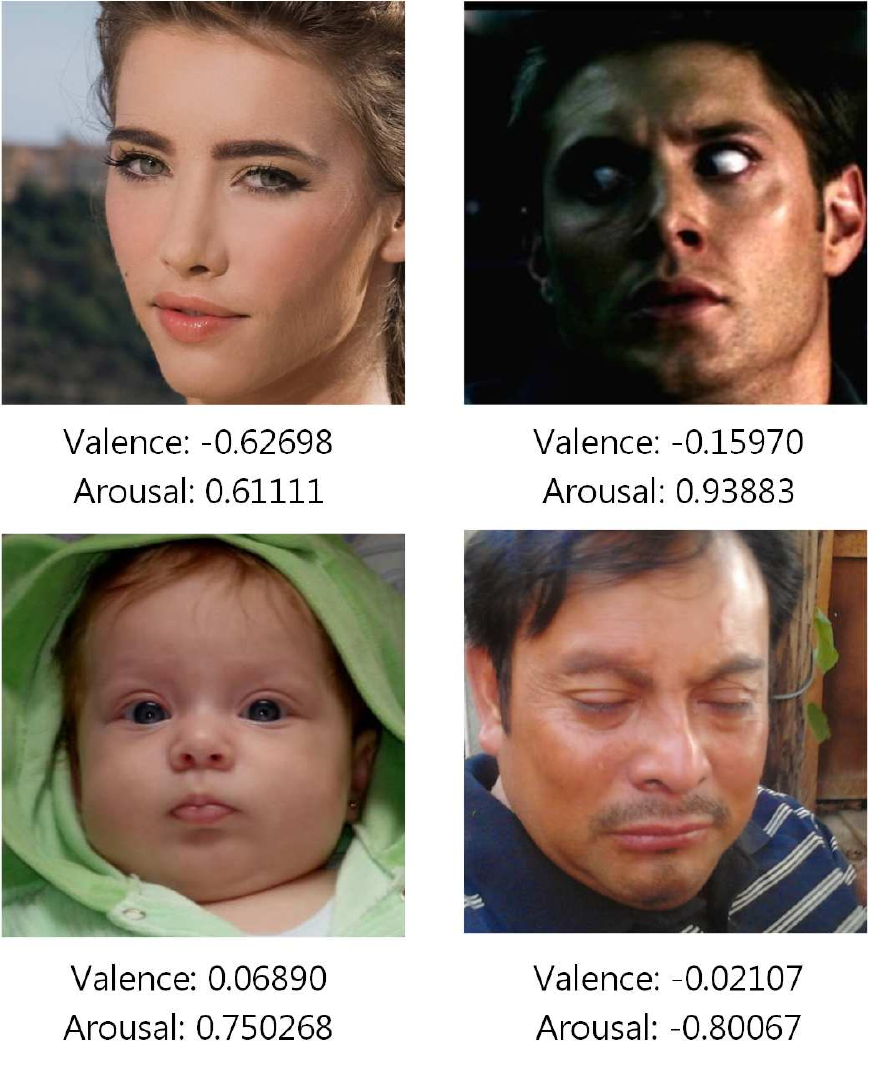}
\caption{Partial samples and their labels of AffectNet.} \label{fig4}
\end{figure}

{\bf AffectNet}: Currently, AffectNet is one of the largest dataset for facial emotion on the discrete and continuous models. The data was crawled under three search engines (Google, Bing and Yahoo) with 1250 emotion-related tags. The collected images is widely distributed in most range of ages. In the dataset, nearly 10\% of the faces have glasses, 50\% of the face have makeup on the eyes and lips and the postures of the face are also various. Therefore, the distribution of the AffectNet is extraordinarily similar to the real world. About 300,000 face images are correctly labeled as continuous values by crowdsourcing. The values of valence and arousal are in the range of [-1,1]. Since the author has not published the test set yet, we use the validation set published by the author to validate the approach proposed in the paper. Figure~\ref{fig4} shows some samples and their labels of AffectNet dataset.

In order to measure the similarities and variations of ground-truths and predictions on dimensional models, there are several metrics, including Root Mean Square Error(RMSE), Concordance Correlation Coefficient(CCC), et al. In this work, we calculate the two metrics to evaluate our proposed framework. In the following, we briefly introduce the definitions of these metrics.

RMSE can heavily weigh the outliers, but does not take into account the correlation between the data, which is defined as follow:
\begin{equation}\label{7}
RMSE=\sqrt{\frac{1}{n}\sum_{i=1}^{n}(y_i-\hat{y_i})^2}.
\end{equation}

CCC measures the disparity of the data while taking into account the covariance of the data. Consequently, CCC is broadly used in various competitions (AVEC, OMG, Aff-Wild), the definition of CCC is as follow:
\begin{equation}\label{8}
CCC=\frac{2s_{y\hat y}}{s_y^2+s_{\hat y}^2+(\bar y- \bar{\hat y})}.
\end{equation}
Where $s_y$ and $s_{\hat y}$ are the variances of the ground labels and predicted labels respectively, $s_{y\hat y}$ represents the covariance of the two. $\bar y$ and $\bar{\hat y}$ are the corresponding mean values.

\subsection{Implements}
In the aspect of image preprocessing, we adopted the current state-of-the-art method, i.e. Multi-task Convolutional Networks (MTCNN)~\cite{zhang2016joint}, to detect facial landmarks and align the faces base these points. The images were cropped and scaled to a size of 56 * 56. After gray scaled the images, we flipped them on the horizontal direction and random cropped to 48*48 as data augmentation. During the test, we did the same processing on the test set, averaging the predicted values of the two flipped images. We exploited Tensorflow to implement our proposed framework and trained it from scratch. For convolution layers and fully connection layers, we initialized with He~\cite{he2015delving} and Xavier~\cite{glorot2010understanding} method, respectively. In order to optimize the parameters in the network, we chose RMSPROP as the optimization algorithm with a batch size 256. The initial learning rate was set to 1e-3, when the loss did not drop, the learning rate was divided by 10. We spent nearly 5 hours to train the entire model for 10 epochs with Titan-X GPU support.

\subsection{Results}
\subsubsection{Baseline}
We explore the advancement of different components proposed in our framework for improved facial emotion recognition. In order to compare our proposed framework and verify the effectiveness of each components, we employ two different neural network models: 2Att-CNN and MultitaskCNN, which both based on the AlexNet[47]. "2Att-CNN" denotes a single-task model which contains two-level attention mechanism while "MulitaskCNN" means that both valence and arousal tasks are performed simultaneously without any other changes. Meanwhile, in the training time, the parameters used in two baselines are consistent with above. This is to ensure that the improvement or diminishment does not stem from hyper-parameters.
\begin{table*}[ht]
\caption{The performance of different level attention mechanisms on AffectNet dataset.}
\centering
\label{tab1}
\begin{tabular}{lllllll}	
\hline
\multicolumn{1}{c}{\multirow{2}{*}{Method}} & \multicolumn{3}{c}{CCC} & \multicolumn{3}{c}{RMSE} \\ \cline{2-7}
\multicolumn{1}{c}{} & \multicolumn{1}{c}{Valence} & Arousal & Mean value & Valence & Arousal & Mean value \\ \hline
MultitaskCNN & 0.589 & 0.423 & 0.506 & 0.373 & 0.408 & 0.391 \\
1st-level & 0.605 & 0.437 & 0.518 & 0.367 & 0.402 & 0.385 \\
2nd-level & 0.618 & 0.460 & 0.539 & 0.360 & 0.393 & 0.377 \\ \hline
\end{tabular}
\end{table*}

\subsubsection{Two-level attention mechanism learning}
In this section, we conduct extensive experiments on a open dataset to validate the effectiveness of two-level attention approach compare to baseline. In the phase of face detection and alignment, partial samples cannot be located due to large occlusions or deflection angle. Owing to the large amount of data in the AffectNet, the failed samples both participate in training and testing. For the fairness, the results on AffectNet are derived from the mean of five experiments with same parameters.

Table~\ref{tab1} shows the experiment results of employing different levels of attention mechanisms. "1st-level" denotes adopting first-level attention mechanism to "MulitaskCNN", and "2nd-level" represents adopting two-level attention mechanism to "MulitaskCNN". Compared with the results of "MulitaskCNN", the performance of adding 1st-level attention mechanism has improved slightly both on CCC and RMSE. The experimental result on two-level attention mechanism is better than that of "MulitaskCNN" and "1st-level" model. From the comparison of Table~\ref{tab1}, we can see that the soft weight attention of mask branch in "1st-level" model boosted the quality of the extracted features. The improvement of "2nd-level" based on "1st-level" model also enhance the effect, which demonstrate the effectiveness of "2nd-level" model.

We also explore the impact of the number of attention block in first-level attention on the experimental result. All experiments report in Table~\ref{tab3} use two-level attention mechanism. The only difference is the number of blocks in the first-level attention. It is clear that the model with two attention blocks yielded the best result. The performance does not increase as the number of attention blocks increase. The use of a one-block only weights the first-level attention features but weakens the role of 2nd-level attention mechanism. For the 3-block model, due to the small size of data in the experiment, the effect of 1st-level attention on the small feature map of high-level is not obvious. And the increase in the number of blocks also deepens the depth of network, resulting in poor learning of low-level features. Thus, we adopt the 2-block model to conduct other experiments.
\begin{table*}[ht]
\caption{The performance using different residual attention blocks on AffectNet dataset.}
\centering
\label{tab3}
\begin{tabular}{lllllll}	
\hline
\multicolumn{1}{c}{\multirow{2}{*}{Method}} & \multicolumn{3}{c}{CCC} & \multicolumn{3}{c}{RMSE} \\ \cline{2-7}
\multicolumn{1}{c}{} & \multicolumn{1}{c}{Valence} & Arousal & Mean value & Valence & Arousal & Mean value \\ \hline
1-Block & 0.677 & 0.511 & 0.594 & 0.365 & 0.382 & 0.373\\
2-Block & 0.714 & 0.556 & 0.635 & 0.353 & 0.364 & 0.359 \\
3-Block & 0.696 & 0.532 & 0.614 & 0.356 & 0.371 & 0.364\\ \hline
\end{tabular}
\end{table*}

\begin{table*}[ht]
\caption{The performance of our framework and other methods on AffectNet dataset.}
\label{tab4}
\centering
\begin{tabular}{lllllll}	
\hline
\multicolumn{1}{c}{\multirow{2}{*}{Method}} & \multicolumn{3}{c}{CCC} & \multicolumn{3}{c}{RMSE} \\ \cline{2-7}
\multicolumn{1}{c}{} & \multicolumn{1}{c}{Valence} & Arousal & Mean value & Valence & Arousal & Mean value \\ \hline
SVR~\cite{mollahosseiniaffectnet}& 0.372 & 0.182 & 0.277 & 0.384 & 0.513 & 0.449 \\
CNN~\cite{affectnet2018} & 0.600 & 0.340 & 0.470 & 0.370 & 0.410 & 0.390 \\ \hline
2Att-CNN & 0.609 & 0.365 & 0.487 & 0.368 &  0.406 & 0.387 \\
2Att-Mt & 0.618 & 0.460 & 0.539 & 0.360 & 0.393 & 0.377 \\
2Att-2Mt & 0.714 & 0.556 & 0.635 & 0.353 & 0.364 & 0.359 \\ \hline
\end{tabular}
\end{table*}

\subsubsection{Two-stage multi-task structure}
In this section, we conduct experiments to investigate the effectiveness of different multi-task learning structures on AffectNet. Table~\ref{tab4} shows the detailed results for different structure. "2Att-Mt" denotes only utilizing the second-stage multi-task learning, i.e. common multi-task learning which only leverage dimensional representations to predict the value of valence and arousal. "2Att-2Mt" denotes our proposed multi-task learning structure (use multi-task learning to learn categorical and dimensional representations, and then predict the targets). We can easily draw conclusion that the CCC and RMSE performance without multi-task learning was significantly worse than multi-task learning structure. The main reason is that "BasicCNN" does not take advantage of the correlation of different tasks. When applying a common multi-task learning structure, the performance had been improved. Compare with "2Att-Mt", "2Att-2Mt" outperformed it with a large margin both on CCC and RMSE, since dimensional model have always had shortcomings that are not precise enough. From the advancement of performance in Table~\ref{tab4}, we can conclude that categorical representations compensate for defects in dimensional representations and enhance the diversity of features through multi-task learning. These are beneficial to the prediction of targets for facial emotion recognition.

In addition, we examine the set of various numbers of hidden units of categorical and dimensional representations. The number of hidden units and the specific results are shown in Table~\ref{tab5}. It can be seen that the best results have been obtained when dimensional layer consisted of 256 neurons, categorical layer consisted of 128 neurons. This experiment also illustrate from the side that categorical representations play a role in predicting the value of valence and arousal, but not the main role.
\begin{table}
\caption{The performance with different number of hidden units of categorical and dimensional representations on AffectNet dataset.}
\centering
\label{tab5}
\begin{tabular}{lllll}
\hline
Dimensional & Categorical & Valence & Arousal & Mean value\\ \hline
 128& 128 & 0.708 & 0.536 & 0.622\\
 256& 128 & 0.714 & 0.556 & 0.635 \\
 256& 256 & 0.711 & 0.549 & 0.630 \\ \hline
\end{tabular}
\end{table}
\subsubsection{Regression learning}
Since there are some samples that failed in face detection, we cannot do manual processing for every failure images. We adopt Tukey's biweight function to suppress the influence of these samples. Table~\ref{tab6} reports the CCC of regression of our proposed framework with different loss function on AffectNet. As shown in Table~\ref{tab6}, our proposal with Tukey's biweight loss outperform the proposal with MSE loss. The performance validate the effectiveness and robustness of Tukey's biweight loss function on such dataset.
\begin{table}
\caption{Comparisons with different regression methods on AffectNet dataset.}
\centering
\label{tab6}
\begin{tabular}{llll}
\hline
Regression methods & valence & arousal & Mean value\\ \hline
MSE & 0.709 & 0.541 & 0.625 \\
Tukey's biweight & 0.714 & 0.556 & 0.635 \\ \hline
\end{tabular}
\end{table}
\subsubsection{Framework performance}
In the last section, we verify our proposed framework on the validation set of AffectNet. For fair comparison, we use the same training set and validation set as in [20]. Table~\ref{tab4} shows the result of different methods. It can be seen that CCC of arousal on our model (both "2Att-Mt" and "2Att-2Mt") is improved enormously than others. The performance of valence have not been greatly improved like arousal. Since we found that the loss of valence decreased expeditiously in the beginning of training and quickly converged, while the loss of arousal declined slightly. The results of both SVR and CNN also demonstrated that the features of valence are effortlessly to learned, but the features of arousal are arduously to learn. The results also prove that our proposed framework can effectively extract attention-grabbing features, utilize the correlation of categorial and dimension model and the correlation of two targets. In addition, the input size of our framework is 48*48, which is one twenty-eighth of that in~\cite{mollahosseiniaffectnet}, which also evaluate that our approach is capable to achieve acceptable results with lower resolution images.
\begin{table}[h]
\caption{The performance of facial emotion classification with different methods on AffectNet dataset.}
\centering
\label{tab7}
\setlength{\tabcolsep}{0.8mm}{
\begin{tabular}{ll}
\hline
 Methods & Accuracy \\
Down-sampling & 0.50  \\
Up-sampling & 0.47 \\
Weighted-loss & 0.58 \\ \hline
Ours & 0.48 \\ \hline
\end{tabular}}
\end{table}
In order to verify the generalization ability of our proposed framework, we employ our trained model to classify the AffectNet to seven categories directly, without any change. Table~\ref{tab7} shows the result for facial expression classification. We can observe that the difference between ours and "down-sampling" in~\cite{affectnet2018} is relatively small, even higher than the result of "up-sampling". According to the result in the fourth column, it cleverly handles data imbalance. Although there is a serious imbalance in the data, and the model we designed is mainly for the estimation of continuous dimensions, we still got a good result on the categorical model.

\subsection{Conclusion}
\label{sec5}
In this paper, we constructed a two-level attention with two-stage multi-task learning framework for facial emotion estimation. In order to enhance the ability of neural network to extract features, we employed residual attention block to extract position-level features automatically. And Bi-RNN with self-attention was proposed to seize the relationship between layer-level features adaptively. And then, we designed a two-stage multi-task learning structure to predict the targets of valence and arousal simultaneously. On the one hand, the structure utilize the categorical representations to compensate for the continuous representations and improve the predicted result. On the other hand, it also leverage the correlation of valence and arousal through multi-task learning. Finally, we adopted Tukey's biweight loss function to suppress the influence of incorrect samples. The performance of our proposed framework on the AffectNet outperformed other methods with a large margin, which proved the superiority of our proposed framework. In addition, we conducted extensively comparative experiments to demonstrate the effectiveness of each components.  We also carried out an experiment on facial expression classification. The result indicated the generalizability of our methods.

However, we only employed AffectNet as the training set, and currently does not apply the datasets that only contain category labels in our work. In the future, we plan to implement multi-domain strategy to our framework, so that our framework can train with more datasets.

\section*{Acknowledgments}
This research has been partially supported by National Natural Science Foundation of China under Grant No.61672202, No.61432004 and NSFCShenzhen Joint Foundation(Key Project)(Grant No.U1613217), and Open Foundation of the Laboratory for Internet of Things and Mobile Internet Technology of Jiangsu Province, Huaiyin Institute of Technology(JSWLW-2017-017).

\bibliographystyle{model1-num-names}
\bibliography{Two-level Attention with Two-stage Multi-task Learning for Facial Emotion Recognition}

\end{document}